\documentclass[letterpaper, 10 pt, journal, twoside]{ieeetran}

\IEEEoverridecommandlockouts                              % This command is only needed if 
\usepackage{graphics} % for pdf, bitmapped graphics files
\usepackage{subfiles}
\usepackage{csquotes}
\usepackage{gensymb}
\usepackage{pifont}
\usepackage{textcomp}
\usepackage[table]{xcolor}
\usepackage{todonotes}
\usepackage{pdfpages}
\usepackage{blindtext}
\usepackage{soul}
\usepackage{hyperref}
\usepackage{graphicx}
\usepackage{mathtools}
\usepackage{multirow}
\usepackage{booktabs}
\usepackage{cite}
\usepackage{flushend}
\usepackage{makecell}

\usepackage{amsfonts}
\usepackage{algorithm}
\usepackage[inline]{enumitem}
\usepackage[noend]{algpseudocode}
\usepackage{caption}
\graphicspath{{images/}{../images/}}
\usepackage{subcaption}
\usepackage[font={footnotesize}, skip=5pt]{caption}
\usepackage[export]{adjustbox}
\usepackage[resetlabels]{multibib}
\newcites{New}{References}
\usepackage{cuted}
\usepackage{amsmath} 
\usepackage{amssymb} 
\usepackage{amsfonts}
\usepackage{bm} 
\usepackage{siunitx}
\usepackage{cleveref}
\usepackage{cancel}
\usepackage[inline]{enumitem}

\usepackage{microtype}

\renewcommand{\eqref}[1]{Eq.~(\ref{#1})}
\newcommand{\figref}[1]{Fig.~\ref{#1}}
\newcommand{\tabref}[1]{Tab.~\ref{#1}}

\usepackage{booktabs}
\usepackage{xspace}
\usepackage{array}
\usepackage{tabularx}
\newcolumntype{Y}{>{\centering\arraybackslash}X}
\newcolumntype{Z}{>{\raggedleft\arraybackslash}X}
\newcolumntype{M}{>{$}c<{$}}

\title{Correct Me if I am Wrong: Interactive Learning for\\Robotic Manipulation}

\author{Eugenio Chisari, Tim Welschehold, Joschka Boedecker, Wolfram Burgard, and Abhinav Valada% <-this % stops a space
\thanks{Manuscript received: October, 7, 2021; Revised December, 19, 2021; Accepted January, 11, 2022.}%Use only for final RAL version
\thanks{This paper was recommended for publication by Editor M. Vincze upon evaluation of the Associate Editor and Reviewers' comments.
This work was supported by the BrainLinks-BrainTools center of the University of Freiburg.} %Use only for final RAL version
\thanks{All authors are with the Department of Computer Science, University of Freiburg, Germany. {\tt\footnotesize chisari@cs.uni-freiburg.de}}
\thanks{Digital Object Identifier (DOI): see top of this page.}
\thanks{© 2022 IEEE.  Personal use of this material is permitted.  Permission from IEEE must be obtained for all other uses, in any current or future media, including reprinting/republishing this material for advertising or promotional purposes, creating new collective works, for resale or redistribution to servers or lists, or reuse of any copyrighted component of this work in other works.}
}

\begin{document}

\maketitle

% Paper headers
\markboth{IEEE Robotics and Automation Letters. Preprint Version. Accepted January, 2022}
{Chisari \MakeLowercase{\textit{et al.}}: Correct Me if I am Wrong: Interactive Learning for Robotic Manipulation} 
% Use only for final RAL version

%%%%%%%%%%%%%%%%%%%%%%%%%%%%%%%%%%%%%%%%%%%%%%%%%%%%%%%%%%%%%%%%%%%%%%%%%%%%%%%%
\begin{abstract}
Learning to solve complex manipulation tasks from visual observations is a dominant challenge for real-world robot learning. Although deep reinforcement learning algorithms have recently demonstrated impressive results in this context, they still require an impractical amount of time-consuming trial-and-error iterations. In this work, we consider the promising alternative paradigm of interactive learning in which a human teacher provides feedback to the policy during execution, as opposed to imitation learning where a pre-collected dataset of perfect demonstrations is used. Our proposed CEILing (Corrective and Evaluative Interactive Learning) framework combines both corrective and evaluative feedback from the teacher to train a stochastic policy in an asynchronous manner, and employs a dedicated mechanism to trade off human corrections with the robot's own experience. We present results obtained with our framework in extensive simulation and real-world experiments to demonstrate that CEILing can effectively solve complex robot manipulation tasks directly from raw images in less than one hour of real-world training.\looseness=-1
\end{abstract}
% Keywords appear just beneath the abstract. Use only for final RAL version. 

%%%%%%%%%%% CONTENT OF THE PAPER %%%%%%%%%%%
\section{Introduction} % Aim for ~1 page
\label{sec:intro}

% Establish the task or problem to solve and why it is hard
%In this work, we consider the problem of training an autonomous agent to solve complex manipulation tasks directly from image observations via feedback from a human teacher. 
% Talk about the state of the art. Where do they shine and where do they struggle?
\IEEEPARstart{D}{eep} reinforcement learning (RL) approaches have achieved tremendous success in many real-world robotic tasks~\cite{haarnoja2018learning, hwangbo2019learning, zeng2020tossingbot, chisari2021learning, honerkamp2021learning, burgard2020perspectives}. However, their advancement has been limited to domains in which a simulator is available or to environments that have been tailored and instrumented for the agent's training. This can be attributed to the fact that standard RL algorithms suffer from one or more of the following issues~\cite{dulac2019challenges}, which hinders their applicability to real-world scenarios. First, they have high sample complexity due to which millions of training samples are required to successfully learn to solve a task. Second, they require engineering a reward function. On one hand, dense reward functions are prone to suffer from unexpected exploitation strategies by the agent. On the other hand, sparse reward functions provide a weak learning signal to the agent, resulting in hard exploration and credit assignment problems. In most cases, instrumentation of the environment is also required to provide the reward signal to the agent and to reset the environment after each episode. Lastly, untrained autonomous agents are likely to undertake dangerous and unsafe behavior while optimizing their policy. This is in stark contrast to how we humans learn most skills since we usually do not need to engage in extensive trial and errors to learn new tasks. As a basic example, people do not need to crash multiple times to learn how to drive.

\begin{figure}
  \centering
  \includegraphics[width=\linewidth]{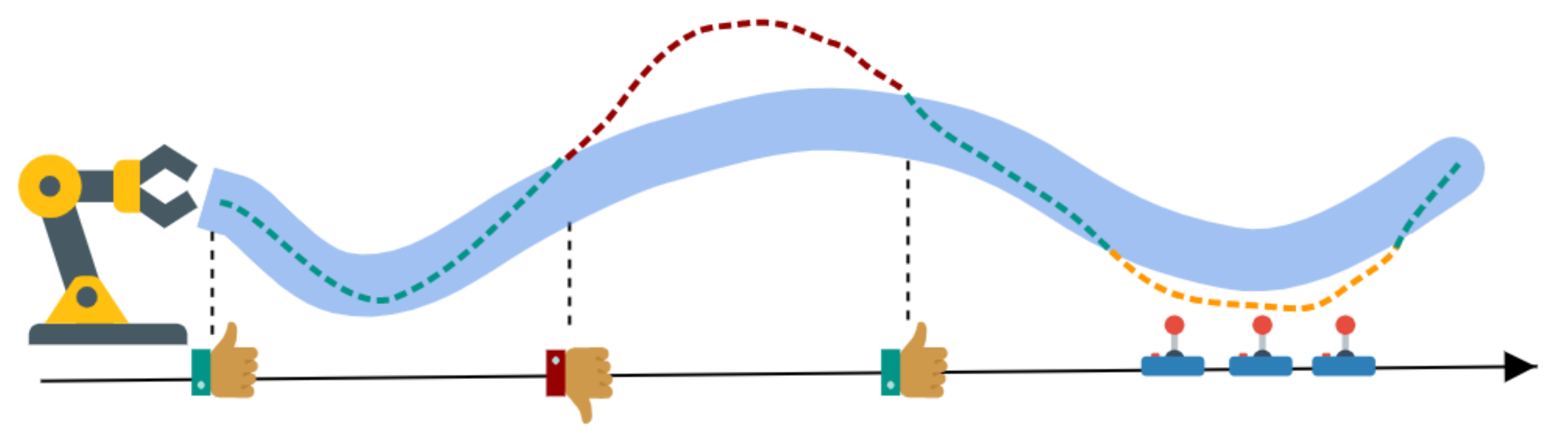}
  \caption{The two types of feedback that the robot receives from the teacher. At the start of the episode, we assume a positive evaluative feedback \( q=+1 \). If the robot trajectory becomes unsatisfactory and the human teacher is not able to easily correct it, he/she can toggle the evaluative feedback to \( q=0 \). All subsequent states will be also labeled as zero, until \( q \) is toggled back to positive. Alternatively, if the path can be easily corrected by acting on one or two degrees of freedom, the teacher can do so, in which case the label is set to \(q=\alpha\).}
  \label{fig:feedback}
%   \vspace{-0.4cm}
\end{figure}

The main difference lies in the fact that humans can leverage useful prior knowledge instead of learning from scratch. In order to gain such prior knowledge, we usually rely on more experienced individuals such as parents and teachers. Skills such as speaking, writing, math, playing a sport, or tying shoelaces, are all learned under the help and supervision of a teacher. Imitation learning (or learning from demonstrations)~\cite{osa2018algorithmic, abbeel2010autonomous, osa2014online, zucker2011optimization, hurtado2021learning} enables exploiting knowledge of an expert to efficiently learn new tasks. Here the expert demonstrates one or more successful trajectories which can then be used by the agent to accelerate learning. However, collecting sufficient demonstrations is often a lengthy and expensive process, and it is not clear a-priori how many demonstrations are needed to train a good policy, often leading to multiple iterations of data collection, re-training and re-evaluation. Additionally, training a policy on expert trajectories suffers from compounding errors and data mismatch issues since sequential decision problems violate the common i.i.d. assumption~\cite{ross2011reduction}.

% Tease your idea with few sentences and a "pull" figure
In this work, we propose to overcome these limitations by adopting an interactive learning approach in which a human teacher provides feedback to the robot during training. The interactive learning setting has three main advantages compared to the learning from demonstration framework: 
\begin{enumerate*}[label=(\arabic*)]
\item the distribution of data collected is induced by the policy itself, rather than being provided a-priori by an expert, avoiding data mismatch issues,
\item it allows the agent to learn self-correcting behaviors, i.e. to learn how to recognize a failed attempt and correct accordingly.
\item it allows the human teacher to directly see the improvement in the policy performance, facilitating the decision about when to stop training.
\end{enumerate*}
In particular, we combine both evaluative and corrective feedback from the teacher to train the robot as illustrated in \figref{fig:feedback}. We call this framework Corrective and Evaluative Interactive Learning (CEILing). We show that our approach addresses many of the challenges required for real-world robotic applications such as those delineated in~\cite{dulac2019challenges}. CEILing is able to learn directly on real robots from limited samples within an hour, and to deal with high-dimensional continuous state (raw images) and continuous action space (seven-dimensional) while acting in real-time at a frequency of 20~Hz. Our method does not require any reward function, thus avoiding credit assignment and reward exploitation issues altogether. The presence of the human teacher makes sure that the robot can be stopped in the case of unsafe behavior, thus avoiding the risk of causing any damage. 

In summary, our contributions are the following:
\begin{enumerate}
\item We introduce CEILing, a novel human-in-the-loop system for training robot policies which combines both evaluative and corrective feedback from a human teacher.
\item We conduct an extensive user study with 12 different participants to evaluate the performance of CEILing compared to multiple baselines when learning from untrained users.
\item We present real world experiments which show that CEILing allows to train a physical robot to solve complex manipulation tasks from raw monocular images in less than one hour.
\item We make the code, models and videos publicly available at \url{http://ceiling.cs.uni-freiburg.de}.
\end{enumerate}

%===============================================================================

\section{Related Work} % Aim for ~1 page

The goal of our work is to leverage feedback from a human teacher to effectively train an autonomous agent. The initial approaches that address this problem were inspired by animal clicker training, where human feedback was used as a reward signal in reinforcement learning~\cite{kaplan2002robotic, blumberg2002integrated}. Several methods have been proposed that can be categorized into three main groups according to the type of feedback provided by the human teacher: evaluative, corrective, and comparison feedback.

\subsection{Evaluative feedback}
Evaluative feedback consists of a scalar value that the human provides to the agent indicating the quality of the current agent behavior. In contrast to the reward function of standard RL, such evaluative feedback takes the long-term consequences of each action into account, from the prior domain knowledge of the human teacher. The TAMER framework~\cite{knox2008tamer, knox2009interactively, warnell2018deep} uses this feedback to learn a reinforcement function \( H:S \times A \rightarrow \mathbb{R} \) by supervised learning. The agent then acts greedily with respect to the learned function \( H \). This can be considered equivalent to learning a \(Q^*\) function by regressing the human-provided feedback directly. Similarly, in the policy shaping framework~\cite{griffith2013policy, cederborg2015policy}, the human feedback is considered as a label on the optimality of the action rather than a reward. Additionally, \cite{macglashan2017interactive} and \cite{arumugam2019deep} propose the framework of convergent actor critic by humans in which the human feedback is considered as the advantage function \(A^{\pi}(s, a) = Q^{\pi}(s, a) - V^{\pi}(s) \) rather than the optimal \(Q^*(s, a)\) function, and instead of using this feedback to learn an approximated \(\hat{A}^{\pi}(s, a)\), it is directly applied to a policy gradient update rule. 

Instead of labelling each state-action pair individually, \cite{cui2018active} propose an active reward learning from critique approach, where the teacher evaluates the robot behavior by marking segmentation points along the generated trajectory. Similarly to our work, this allows the user to understand actions in the context of a trajectory and to collect many state action labels from a small number of segmentation points. Nevertheless, marking a pre-generated trajectory requires some type of interface to visualize the trajectory and mark the segmentation points. On the other hand, in our work the segmentation points are marked “on policy”, i.e. while the robot executes the trajectory, which does not require any specific interface other than the press of a button at the right time.

\subsection{Corrective feedback}
Corrective feedback is used to directly provide information to the agent about how to improve its actions. \cite{chernova2009interactive} and  \cite{mericcli2010complementary} introduced the idea of corrective demonstrations where the human feedback is used to correct bad actions of the agent. The corrected state-action pairs are then added to a dataset which is used to update the policy. Similarly, \cite{kelly2019hg} propose HG-DAgger where the human teacher feedback is used to steer trajectories drifting to an unsafe state space back to a safe set. Compared to the classical imitation learning algorithm DAgger~\cite{ross2011reduction}, the human teacher is in charge of deciding when to intervene, rather than being queried by the algorithm. In the similar Corrective Advice Communicated by Humans framework~\cite{celemin2015coach, perez2019continuous}, the human provides only a direction of improvement of the action rather than an optimal value. 

The mentioned works only use the expert labeled actions for training and discard all other state-action pairs. On the other hand, \cite{mandlekar2020human} recently showed that also including the non-corrected portion of each trajectory is helpful to reinforce the good behavior and improve the robustness of the policy. Since the dataset usually contains an imbalanced amount of non-corrected and corrected samples, their proposed Intervention Weighted Regression (IWR) method uses a weighting parameter \(\alpha\) which correspond to the amount of prioritization given to the intervention samples. Nevertheless, this approach works under the assumption that the teacher is always able to correct bad behaviors. This might not be true in general, since non-expert users might be in charge of training the robot. In our work, the use of evaluative feedback on non-corrected portions of the trajectory gives the human teacher the option to decide which part of the trajectory to use for training and which to discard.
Additionally, Mandlekar et al.~\cite{mandlekar2020human} use groundtruth low-level state observations such as object poses and binary indicator for contact, which are usually only available in simulation. In our work on the other hand, high-dimensional image observations are used, which allow us to apply the method to a physical robot.\looseness=-1

\subsection{Comparison feedback}
Comparison feedback, also called preference-based learning, involves communicating which of the different trajectories is preferable to the agent. \cite{akrour2011preference} and \cite{schoenauer2014programming} propose to combine preference-based learning with active ranking for discrete as well as continuous benchmark problems. \cite{jain2013learning,jain2015learning} present a co-active online learning framework where the human user iteratively provides slight improvements over the trajectory currently proposed by the system. \cite{palan2019learning} and \cite{biyik2019asking} introduce the DemPerf framework where demonstrations and comparison feedback are used to learn the weights of a linear reward function which is then used to solve robotic manipulation problems. Later, this idea was extended by modeling the reward function as a Gaussian Process~\cite{biyik2020active}. The aforementioned works demonstrate how robot trajectories can be improved by a human teacher providing comparison feedback, but they were limited to low dimensional states as inputs. One exception is~\cite{christiano2017deep} where the authors employ deep neural networks to deal with high dimensional observations. Nevertheless, the approach is only applicable for simulated environments, since trajectories need to be re-sampled and replayed on screen for the user to provide feedback. Moreover, it requires millions of steps to converge to a good solution. The method we propose in this work instead learns from high dimensional observations from a human teacher in real-time, requiring only less than one hour of real-world training.

\section{Technical Approach} % Aim for a nice system figure at the beginning + a technical section with some math + a pseudocode algo 
In this section, we introduce CEILing (Corrective and Evaluative Interactive Learning). CEILing combines both corrective and evaluative feedback from a human teacher to train an autonomous agent to effectively solve complex robotic manipulation tasks. The observation space consists of raw image data from a wrist mounted camera as well as the current robot joint and gripper states, while the action space consists of the desired end-effector pose change (6-dimensional) plus the gripper opening and closing action (1-dimensional). The policy does not make use of any pre-trained component as visual features encoder and can be trained on a single GPU. Thus, CEILing can be applied to robots in the wild, without the need for any environment instrumentation.

\subsection{Policy Model}

\begin{figure}
  \centering
  \includegraphics[width=\linewidth]{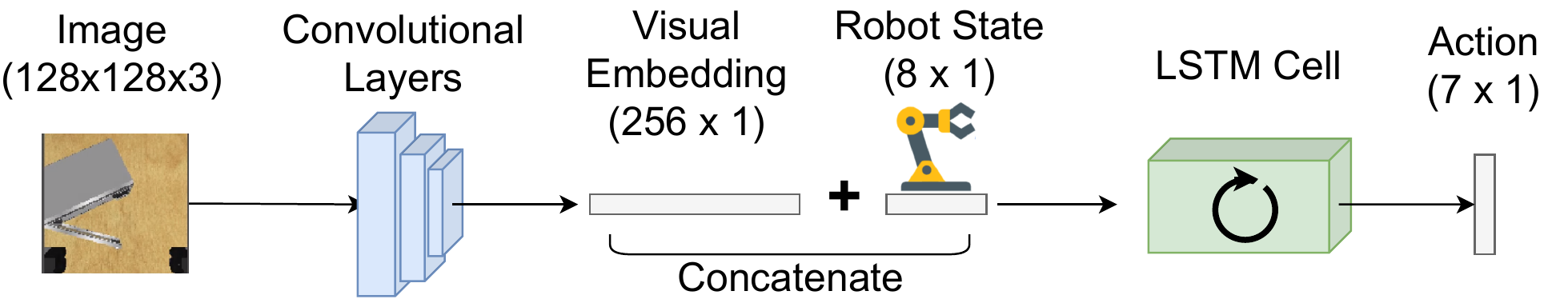}
  \caption{Diagram of the model architecture. First, the image from the wrist mounted camera is fed through 3 convolutional layers with ELU activation function. The output visual embedding is flattened and concatenated with the robot joints and gripper state. They are then fed into an LSTM cell, which outputs the desired action.}
  \label{fig:network}
%   \vspace{-0.4cm}
\end{figure}

In this work, we consider a stochastic policy \( \pi_\theta(a|s) \), which represent the probability distribution of the action \(a\) given a state \(s\). It is parametrized as a Gaussian distribution whose mean is the output of a neural network \(f_\theta(s, \theta)\) with weights \( \theta \) and whose variance is a state-independent parameter \(\sigma ^ 2\):
\(
\pi_\theta(a|s) \sim \mathcal{N}(f_\theta(s, \theta); \sigma ^ 2)
\).
A narrow distribution leads to high losses even for small errors, and therefore a more aggressive optimization, whereas a wider distribution leads to better robustness against noise in the data. Given the scale and speed of our robot, we choose \( \sigma \) to correspond to 1 mm and 0.25 degrees for the translational and rotational components of the action respectively. \figref{fig:network} shows the network architecture which consists of different building blocks. First, the input image is fed through 3 convolutional layers with ELU activation function, using a kernel size of 3 and a stride of 2. The resulting low-dimensional visual embedding is then concatenated with the current robot joint and gripper state, and subsequently fed to an LSTM cell to exploit the sequential data. Finally, the output is mapped to the right action dimension. The model is trained using the Adam optimizer with a learning rate of \( 3 \cdot 10^{-4}\) and weight decay of \( 3· \cdot 10^{-6}\).

\subsection{Interactive Learning}
\label{sec:data_collection}

\begin{algorithm}
\footnotesize
  \caption{CEILing: Corrective and Evaluative Interactive Learning}
  \label{alg:pseudocode}
  \begin{algorithmic}[1]
    \Require replay buffer $\mathcal{D}$, policy $\pi_\theta(a \vert s)$, learning rate $\lambda$
    
    \Statex
    \Function{Main}{}
    \State $\mathcal{D} \gets $ \Call{LoadDemonstrations}{10}
    \Comment{Fill the replay buffer}
    \State \Call{UpdateLoop}{ }
    \Comment{Start Update Loop, asynchronously}
    \State \Call{Sleep}{120}
    \Comment{Train on the demonstrations for 2 minutes}
    \State \Call{EnvironmentLoop}{200}
    \Comment{Run for 200 episodes}
    \EndFunction
    
    \Statex
    \Function{UpdateLoop}{ }
      \While{not done}
        \State $(s, a, q)$ $\gets$ \Call{Sample}{$\mathcal{D}$} 
        \Comment{Sample a batch of data from $\mathcal{D}$}
        \State $L(s, a) \gets -q\log(\pi_\theta(a \vert s))$
        \Comment{Do a forward pass}
        \State $\theta \gets \theta - \lambda \nabla_\theta L(s, a)$
        \Comment{Update policy weights}
      \EndWhile
    \EndFunction
    
    \Statex
    \Function{EnvironmentLoop}{episodes}
      \ForAll{episodes}
      \State success $\gets 0$
      \State $\Call{TaskReset}{}$
      \Comment{Reset environment}
      \While{not (success or timeout)}
        \State $s \gets $ \Call{GetState}{}
        \Comment{Get joint states and camera image}
        \State $a \gets$ \Call{Mean}{$\pi_\theta(a \vert s)$}
        \Comment{Use policy to predict action}
        \State $ a \gets a +$ \Call{GetCorrection}{} 
        \Comment{Teacher Correction}
        \State success $\gets$ \Call{Step}{a}
        \Comment{Take environment step}
        \State $q \gets$ \Call{GetEvaluation}{}
        \Comment{Teacher evaluation}
        \State $\mathcal{D} \gets \mathcal{D} \cup \{s, a, q\}$
        \Comment{Fill replay buffer}
      \EndWhile
      \EndFor
    \EndFunction
  \end{algorithmic}
\end{algorithm}

CEILing needs to operate in a severe low-data regime since the human teacher cannot be expected to supervise the robot training for hours or days. We therefore store all of the available data in a replay buffer. Given the random behavior of an untrained policy, the human teacher would be required to intervene most if not all of the time during the first few episodes, making this phase not much different from collecting actual demonstrations. Hence, we prefer to collect $10$ demonstrations to warm start the policy by regressing on these demonstrations. After the warm start phase, the interactive learning phase starts where the robot applies the latest version of the policy to generate actions at a frequency of 20~Hz. 

At the beginning of each episode, we start with a positive evaluative feedback label \(q=+1\). As long as the trajectory remains appropriate, all the subsequent steps will be automatically labeled with this value, which helps reinforce the good behavior of the robot. When the teacher believes that the robot trajectory can no longer be considered satisfactory and it is not easy to correct, the teacher can toggle the evaluative label to \(q=0\). From there on, all subsequent labels are going to be considered as zero, until they are toggled back to positive by the human teacher. This way we can label all state-action pairs without requiring too much effort from the teacher. If the trajectory is not on the right path but it is easily adjustable, the human teacher can provide corrections to the robot. In this case, all state-action pairs will be labeled with \( q=\alpha \). The weight \(\alpha\), inspired from~\cite{mandlekar2020human}, is defined as the current ratio between the amount of non-corrected and corrected samples in the replay buffer, updated throughout the learning process. It is used to increase the prioritization of the corrected samples in order to counteract the imbalance of the collected data. 
In our experiments, we provide both types of feedback with a gamepad, but the same ideas can be applied through different interfaces such as a space mouse or a VR set, depending on the requirements of the application. Evaluative feedback simply consists of pressing a button at the appropriate time, whereas corrective feedback requires the human teacher to intervene and teleoperate the robot.
The overall interactive learning strategy is depicted in \figref{fig:feedback}. In our experiments, we train for 100~episodes, which roughly correspond to 20~minutes of real-world training, in contrast to millions of episodes needed by standard RL algorithms. 

\subsection{Update Strategy}

While the robot interacts in the environment to collect more data, we also continuously update its policy. In order to maximize the number of update steps and not be limited by the number of steps in the environment, we run the policy update loop in a separate asynchronous thread. An asynchronous policy update loop allows to achieve a very high rate of policy updates, which happen continuously over the course of the experiment. Achieving the same amount of updates by training the network only in between episodes would require long interruptions, which is not desired since it would drastically increase the time effort of the human teacher.
In our setting, we did not experience any slow down in terms of control frequency of the robot nor in the latency of the feedback signal because of the asynchronous training loop. In case the computational load becomes excessive, a simple solution would be to assign higher priority to the control loop or even to run the training loop on a separate machine. At each update step, a batch of labeled and potentially corrected trajectories is sampled from the replay buffer. The goal is to reinforce the good behaviors while ignoring the bad ones, or in other words to increase the probability of the policy picking a good action. To achieve this, we resort to a maximum-likelihood objective: we update the policy by minimizing the average of the negative log-likelihood of all state-action pairs  \( (s, a) \) in the batch with respect to the policy distribution, weighted by their label $q$ from the evaluative feedback. Hence, for a batch of \( (s, a, q) \) tuples, the loss function and its gradient are
\begin{align}
L(s, a) &= -q\log(\pi_\theta(a|s)),
\end{align}
\begin{align}
\nabla_\theta L(s, a) &=-q\nabla_\theta\log(\pi_\theta(a|s)).
\end{align}

As noted in~\cite{osa2018algorithmic}, minimizing the negative log-likelihood under a Gaussian distributed policy is equivalent to minimizing the mean squared error loss function commonly used for behavior cloning with deterministic policies. The pseudo-code for CEILing is presented in Algo~\ref{alg:pseudocode}.

\section{Experimental Evaluation} % A results table with lots of numbers + some plots + one additional cute analysis experiment
\label{sec:result}

We design the experiments to show that through interactive learning, it is feasible to learn an end-to-end robot manipulation policy in less than an hour of training. First, we consider the performance of CEILing in a user study with a group of varied participants, and we compare it to five other baselines. This comparison is carried out on four different tasks in a simulation environment for the sake of reproducibility. Finally, we demonstrate the performance of CEILing on a physical robot, again on four different tasks. For each episode in our experiments, the objects in the environment are reset in a random position to demonstrate that the robot is not memorizing a trajectory but is rather learning to generalize. After each interactive training session, we evaluate the performance of the learned policy for 100 episodes, where the policy is exclusively used for inference and the robot does not receive any type of feedback from the human teacher. The most important metric we report is the final success rate achieved during evaluation.

\subsection{User Study Evaluation}

\begin{figure}
  \centering
  \includegraphics[width=\linewidth]{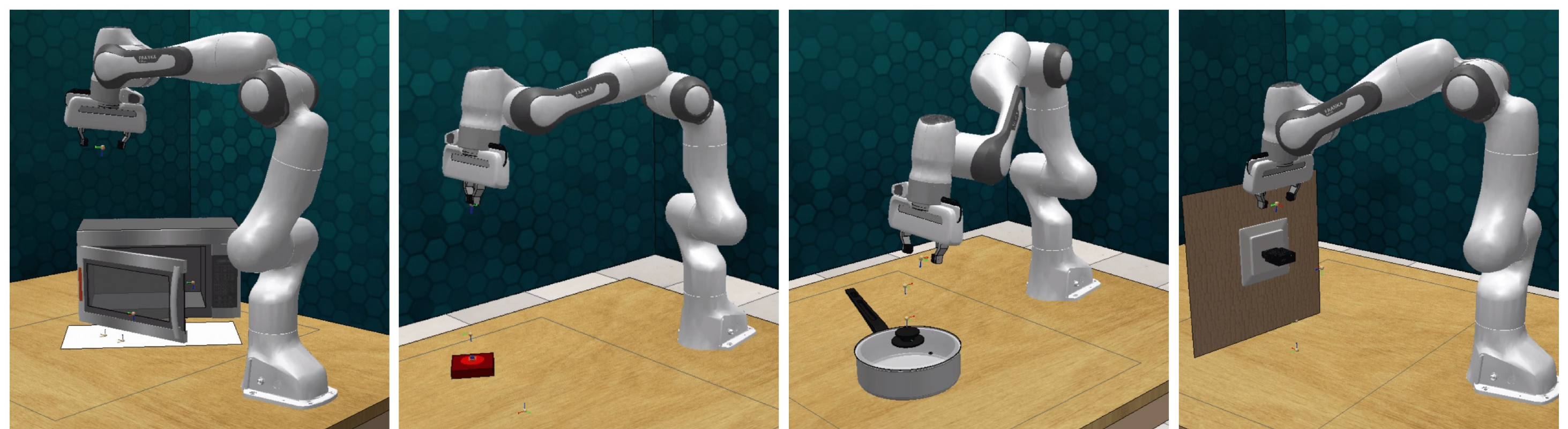}
  \caption{Simulated tasks: CloseMicrowave, PushButton, TakeLidOffSaucepan, UnplugCharger.}
  \label{fig:tasks}
%   \vspace{-0.3cm}
\end{figure}

\begin{figure}
  \centering
  \includegraphics[width=\linewidth]{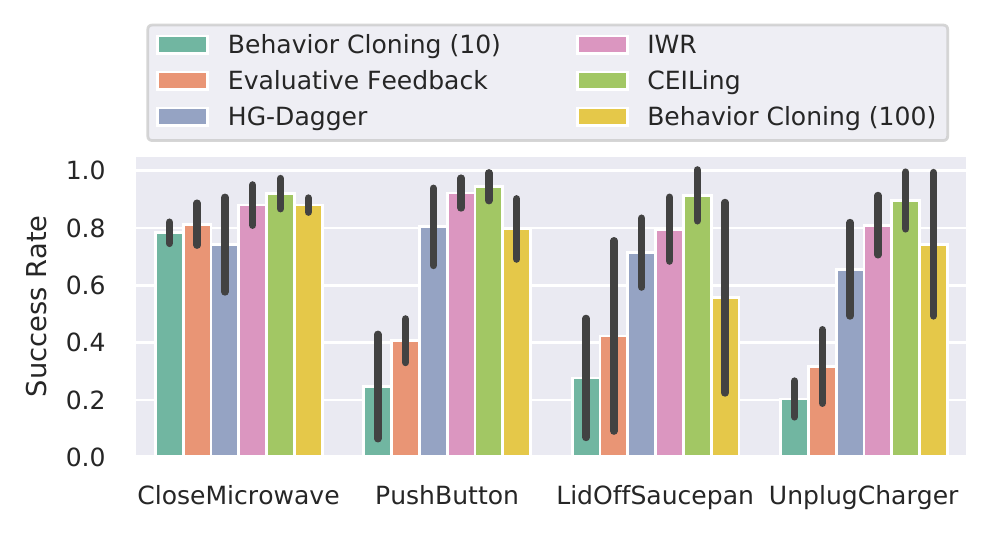}
%   \vspace{-5mm}
  \caption{Comparison of success rates of different policies during evaluation on the four simulated tasks. The average and standard deviation error bars are reported.}
  \label{fig:comparison}
%   \vspace{-0.3cm}
\end{figure}

In this section, we present the results of a user study conducted with a group of 12 participants, both men and women aged between 24 and 32. Each participant was given 5 minutes to familiarize with the controller interface and the robot in a teleoperation setting before the start of the experiments. We conduct these experiments in simulation, in particular using RLBench~\cite{james2019rlbench}, a suite of realistic robot manipulation tasks designed for real-world robot learning. We consider four different tasks, also depicted in \figref{fig:tasks}: CloseMicrowave, PushButton, TakeLidOffSaucepan, UnplugCharger. Our setup consisted of a Franka Emika Panda robot with a wrist mounted camera. Each experiment is repeated three times, each one with a different user and a different random seed. 
Each task was evaluated by $5$ different users, each conducting experiments for all baselines and our proposed approach.

\subsubsection{Baselines}
In this study, we compare the performance of policies trained with different methods: behavior cloning (BC)~\cite{osa2018algorithmic} with 10 and 100 demonstrations, evaluative feedback only, HG-Dagger~\cite{kelly2019hg}, Intervention Weighted Regression (IWR)~\cite{mandlekar2020human} and our proposed method CEILing. Behavior cloning does not use any interaction with the human teacher. The other four methods on the other hand, differ in the type of feedback they receive and in how this feedback is used, as shown in table~\ref{tab:feedback_diff}. 

The Evaluative baseline only receives a binary feedback from the teacher. The samples which are marked as positive are stored in the replay buffer and used to update the policy, while the samples marked as negative are discarded. This allows the teacher to categorize the on-policy behavior of the robot as good or bad, but it does not allow to provide any corrective feedback. The HG-Dagger policy, on the other hand, allows the teacher to provide corrective feedback to the robot. Here, the corrected actions are stored in the replay buffer, while all other samples are discarded. This means that HG-Dagger does not make use of the good non-corrected samples. The IWR baseline also allows the teacher to provide corrective feedback to the robot, but contrary to HG-Dagger, it also stores the non-corrected portion of the trajectory. While the additional data helps improve the robustness of the policy, IWR makes the assumption that the teacher is always able to correct wrong actions, and no data is discarded. This might not be true in general, since non-expert users might be in charge of training the robot. CEILing on the other hand allows the teacher to use evaluative feedback on the non-corrected portions of a trajectory. This means that with CEILing it is possible, besides providing corrections, to both discard wrong actions as well as store the positive ones. 

We also considered two Reinforcement Learning baselines, namely PPO~\cite{schulman2017proximal} and SAC~\cite{haarnoja2018soft} as state-of-the art on-policy and off-policy algorithms respectively, and use the task success as binary reward at the end of the episode. We considered the standard RL problem without the use of any demonstrations. Because of the hard exploration problem resulting from the sparse reward, these policies were not able to solve the tasks.

\begin{table}
\centering
% \renewcommand{\arraystretch}{1.1}
% \begin{tabular}{p{2.8cm}p{1cm}p{1cm}p{1.1cm}}

\begin{tabular}{lcccc}
\toprule
& Evaluative & HG-Dagger & IWR & CEILing \\
\midrule 
\makecell[l]{Teacher can provide \\ corrections}           & -- & \checkmark  & \checkmark  & \checkmark \\
\midrule
\makecell[l]{Non-corrected bad \\ actions are discarded}   & \checkmark  & \checkmark  & -- & \checkmark \\
\midrule
\makecell[l]{Non-corrected good \\ actions are stored}     & \checkmark  & -- & \checkmark  & \checkmark \\
\bottomrule

\end{tabular}

% \vspace{-0.4cm}
\caption{The difference in the type of feedback used by different methods. CEILing is the only method that can provide corrective feedback, discard wrong actions and store good ones.}
\label{tab:feedback_diff}
\end{table}

\subsubsection{Qualitative Results}

\begin{table}
\centering
\setlength{\tabcolsep}{3.8pt}

\renewcommand{\arraystretch}{1.05}
%\begin{tabular}{p{2.8cm}p{1cm}lp{1cm}p{1.1cm}p{1.1cm}}
\begin{tabularx}{\columnwidth}{l M M M M M}
\toprule
& \multicolumn{1}{c}{Evaluation} & & \multicolumn{3}{c}{Training} \\
\cmidrule{2-2} \cmidrule{4-6}
& \text{Success} & & \text{Update} & \text{Duration} & \text{Feedback} \\
& \text{Rate [\%]} & & \text{Steps} [-] & \text{[minutes]} & \text{Rate [\%]} \\
\midrule 
\textbf{CloseMicrowave}\\
Behavior Cloning (10)           & 78 \pm 5 & & 2000 & -- & -- \\
Behavior Cloning (100)          & 88 \pm 3 & & 2000 & -- & -- \\
Evaluative Feedback             & 81 \pm 9 & & 2882 & 22 & -- \\
HG-Dagger~\cite{kelly2019hg}    & 74 \pm 18 & & 2067 & 21 & 25 \\
IWR~\cite{mandlekar2020human}   & 88 \pm 8 & & 1846 & 19 & 18 \\
CEILing (Ours)                  & \mathbf{92} \pm 6 & & 1822 & 18 & \quad\enspace\ 14 \ / \ 2 \\
\midrule 
\textbf{PushButton}\\ 
Behavior Cloning (10)           & 25 \pm 22 & & 2000 & -- & -- \\
Behavior Cloning (100)          & 80 \pm 13 & & 2000 & -- & -- \\
Evaluative Feedback             & 41 \pm 9 & & 2506 & 24 & -- \\
HG-Dagger~\cite{kelly2019hg}    & 80 \pm 15 & & 1729 & 19 & 21 \\
IWR~\cite{mandlekar2020human}   & 92 \pm 6 & & 1574 & 17 & 21 \\
CEILing (Ours)                  & \mathbf{94} \pm 5 & & 1564 & 16 & \quad\enspace\ 21 \ / \ 1 \\
\midrule 
\textbf{TakeLidOffSaucepan}\\ 
Behavior Cloning (10)           & 28 \pm 25 & & 2000 & -- & -- \\
Behavior Cloning (100)          & 56 \pm 41 & & 2000 & -- & -- \\
Evaluative Feedback             & 42 \pm 41 & & 2804 & 29 & -- \\
HG-Dagger~\cite{kelly2019hg}    & 71 \pm 13 & & 2396 & 24 & 17 \\
IWR~\cite{mandlekar2020human}   & 80 \pm 12 & & 2150 & 22 & 21 \\
CEILing (Ours)                  & \mathbf{91} \pm 10 & & 2072 & 21 & \quad\enspace\ 15 \ / \ 7 \\
\midrule 
\textbf{UnplugCharger}\\ 
Behavior Cloning (10)           & 20 \pm 8 & & 2000 & -- & -- \\
Behavior Cloning (100)          & 74 \pm 31 & & 2000 & -- & -- \\
Evaluative Feedback             & 32 \pm 16 & & 2921 & 26 & -- \\
HG-Dagger~\cite{kelly2019hg}    & 66 \pm 18 & & 2032 & 20 & 23 \\
IWR~\cite{mandlekar2020human}   & 81 \pm 12 & & 1927 & 19 & 21 \\
CEILing (Ours)                  & \mathbf{90} \pm 11 & & 1905 & 19 & \quad\enspace\ 16 \ / \ 5 \\
\bottomrule
\end{tabularx}

% \vspace{-0.4cm}
\caption{Results from the simulation experiments, averaged across users. The feedback rate represents the percentage of steps the human teacher had to intervene with corrective feedback. For CEILing, we also show the negative feedback rate, i.e. the percentage of steps the human teacher provided a negative evaluative feedback. The duration is the amount of real time the policy were trained for and update steps are the number of gradient descent updates of the network. The success rate is the percentage of successful episodes during evaluation, and both mean and standard deviation are shown.}
\label{tab:table}
\end{table}

\figref{fig:comparison} shows the average success rate during evaluation as well as standard deviation error bars. Note that for all policies, the same initial pre-training on 10 demonstrations applies. As hypothesized for such a low data regime, behavior cloning trained on 10 demonstrations does not achieve good results. Moreover, interactive learning using only evaluative feedback performs poorly. A combination of small human imprecision in the timing of the feedback together with sub-optimal robot actions provides undesirable state-action pairs to the replay buffer, possibly reducing the quality of the overall training. The next baseline considered is HG-Dagger, which achieves overall higher performance than behavior cloning and pure evaluative feedback. This indicates that corrective feedback provides a stronger learning signal compared to evaluative feedback. We find that the IWR baseline is able to achieve even higher success rate, being consistently above 80\% over all tasks. This confirms the idea that including the on-policy agent samples during training increases the amount of available data and reduces covariate shift, hence improving the overall generalization capabilities of the policy. When we consider the results of CEILing, we see that it achieves overall the highest performance. We can attribute this to the fact that human feedback is inherently prone to mistakes, especially when untrained users are in charge of providing feedback to the robot. Contrary to the other methods, CEILing allows the human teacher to mark and discard the wrong actions that he/she was not able to correct, as well as to store the positive ones.

Lastly, we also compare with a behavior cloning policy trained on 100 perfect demonstrations. Given that collecting perfect demonstrations is a demanding process, this should not be considered an equivalent alternative, but rather an estimate for the achievable performance using the same amount of data in an imitation learning setup. This baseline also has the advantage of training from the beginning on 100 trajectories, in contrast to the interactive learning approaches which start with only 10 trajectories and slowly build up their replay buffer over time, up to 100 episodes. Nevertheless, we observe that CEILing and the IWR baseline achieve overall better performance than this behavior cloning policy. This result highlights the advantages of the interactive learning setting: the distribution of the data collected is induced by the policy itself which drastically reduces the data mismatch and compounding error issues typical of behavior cloning. Namely, expert supervision is added to the dataset where the policy mostly needs it (i.e. where it makes a mistake). 

\subsubsection{Quantitative Analysis}

\tabref{tab:table} shows, besides the success rate, additional metrics for each of these policies, averaged across users.
In particular, the Update Steps column represents the number of updates of the network, the Duration column shows the amount of real time the policy were trained for, and the Correction Rate represents the percentage of steps the human teacher had to intervene with corrective feedback. It is interesting to observe the short training time needed for each run and the low correction rate, which indicates which portion of all the training steps the human teacher had to intervene in with corrections. We find that the success rate does not depend on the amount of corrections, but rather on the quality of the feedback that each teacher is able to provide. Most importantly, this user study highlights the fact that CEILing can be applied by different and untrained users to train a robot's policy and it is able to achieve high performance on all tested tasks, with limited variations among users.

In \figref{fig:users_diff} we present a scatter plot comparing the success rates of each user using our approach and the IWR baseline across all task.
%In order to better evaluate the difference in performance across users between CEILing and IWR, we added an additional analysis of our experimental results in \figref{fig:users_diff}. 
Points lying above the diagonal represent better performance of a user with the CEILing method, whereas points below the diagonal represent better performance with the IWR method. The plot illustrates that most user are able to achieve better results with our proposed method across all tasks.
We observe that less proficient users benefited more from the availability of evaluating feedback, since they were able to discard wrong trajectories when they were not able to provide corrections.

\subsubsection{The Role of Evaluative Feedback}
\begin{figure}
  \centering
  \includegraphics[width=\linewidth]{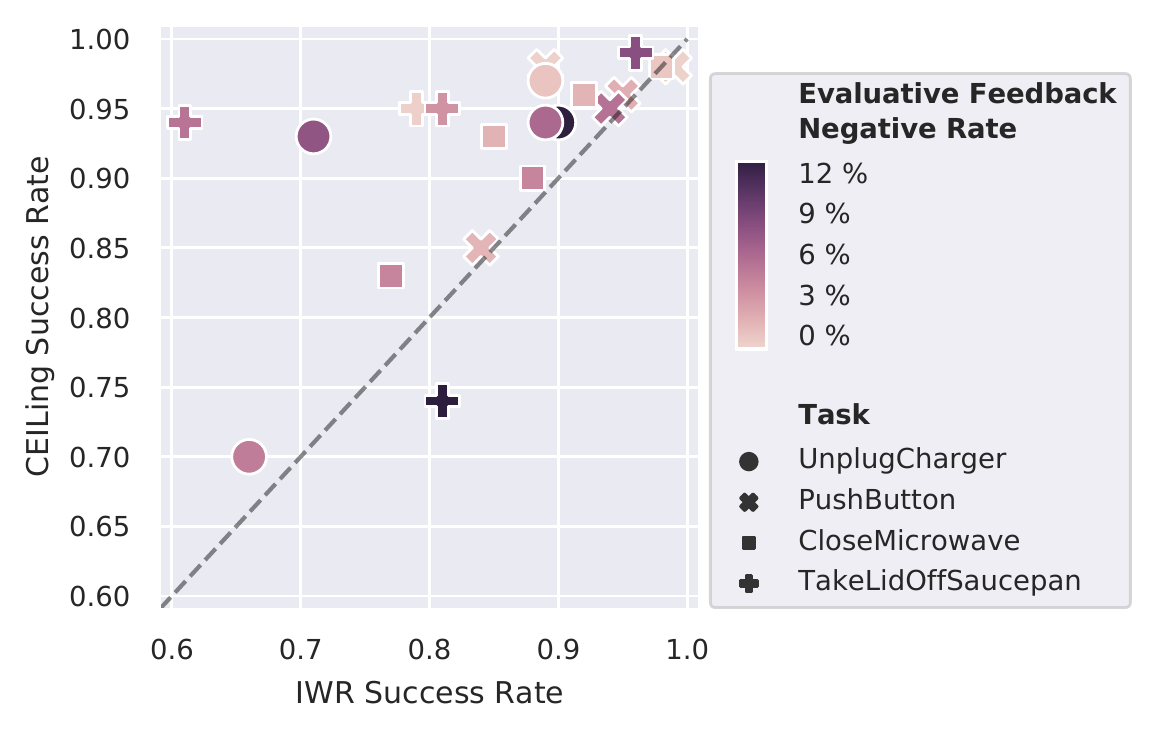}
%   \vspace{-5mm}
  \caption{Scatter plot comparing the success rates of each user using our approach and the IWR baseline across all task, as well as the amount of negative evaluative feedback provided by each user. Points lying above the diagonal represent better performance of a user with the CEILing method, whereas points below the diagonal represent better performance with the IWR method. The plot illustrates that most user are able to achieve better results with our proposed method.}
  \label{fig:users_diff}
%   \vspace{-0.3cm}
\end{figure}

Each participant affirmed, that the option to provide evaluative feedback helps to deal with cases where they are not able to effectively correct the robot. As confirmed in \tabref{tab:table}, the evaluative feedback is used in all tasks.
The benefit of the evaluative feedback stems from a multiple of reasons. The additional complexity of image observations increases the difficulty of the learning process, sometimes leading to unpredictable behavior from the robot during training. In such cases, directly recovering through teleoperation can be too difficult or cognitive demanding, especially for untrained users. Further, as the method is designed to perform manipulation actions, another highly relevant issue arises when the manipulated object is shifted to a irreversible state due to a bad action (e.g pushed from the table). In such situations the subsequent data should not be considered for policy learning. Additionally, although not all degrees of freedom in rotation are required to execute the task successfully, these rotations are not fixed and often are subject to unexpected  changes due to interaction with the environment, i.e., the gripper orientation can change when exerting force on an object in a direction that the object can not move (e.g: pressing the lid of the pan or the microwave door down). Recovering from these unexpected changes is highly demanding and not always successful. At times we also encounter kinematic failures where the desired commands are not executable due to the current joint configuration.  
Other possible situations include moments of distraction or tiredness from the user. In such cases, it is a useful feature for the user to discard the bad behavior, which would otherwise pollute the replay memory.
\figref{fig:users_diff} also shows the amount of negative evaluative feedback provided by each user for the CEILing policy. 
Similar to the corrective feedback, the success rate does not directly correlate to the amount of negative evaluative feedback provided, but rather on the quality of feedback that each user is able to provide.
Moreover, less skilled users who trained worse policies also had to utilize the negative feedback more often, in order to compensate for the higher amount of mistakes from the robot.

\subsection{Real-World Experiments}

\begin{figure}
  \centering
  \includegraphics[width=\linewidth]{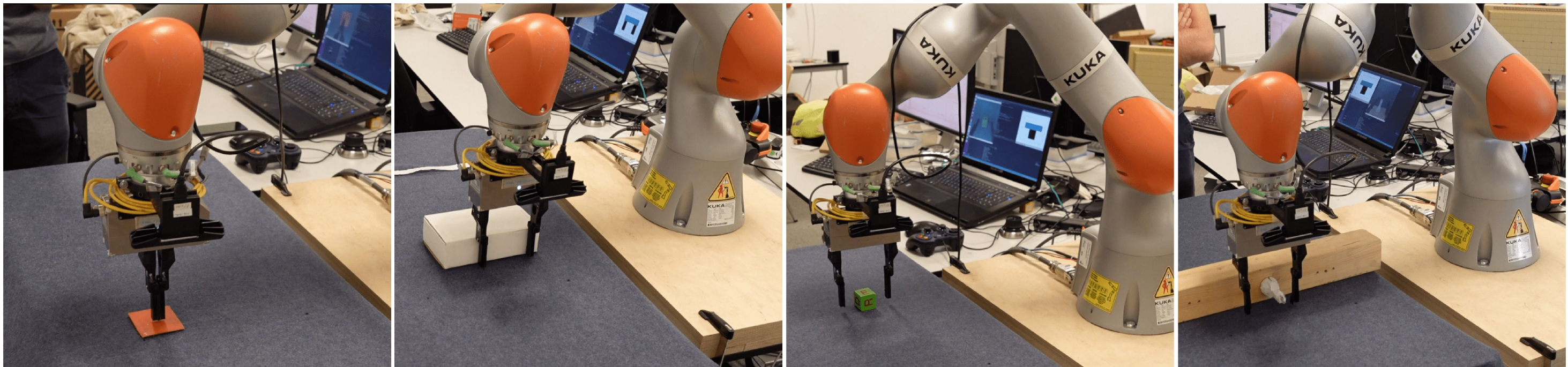}
  \caption{Real robot tasks: Reach, PushBox, PickupCube, PullPlug.}
  \label{fig:kuka_tasks}
\end{figure}

\begin{table}
\centering
\setlength{\tabcolsep}{3.8pt}

\renewcommand{\arraystretch}{1.05}
%\begin{tabular}{p{2.8cm}p{1cm}lp{1cm}p{1.1cm}p{1.1cm}}
\begin{tabularx}{\columnwidth}{l M M M M M}
\toprule
& \multicolumn{1}{c}{Evaluation} & & \multicolumn{3}{c}{Training} \\
\cmidrule{2-2} \cmidrule{4-6}
& \text{Success} & & \text{Update} & \text{Duration} & \text{Feedback} \\
& \text{Rate [\%]} & & \text{Steps [-]} & \text{[minutes]} & \text{Rate [\%]} \\
\midrule 
\textbf{Reach}\\
Behavior Cloning (100)          & 96 & & 4000 & -- & -- \\
HG-Dagger~\cite{kelly2019hg}    & 72 & & 5442 & 33 & 30 \\
IWR~\cite{mandlekar2020human}   & 90 & & 5136 & 34 & 29 \\
CEILing (Ours)                  & \mathbf{98} & & 4326 & 18 & \quad\enspace\ 4 \ / \ 1 \\
\midrule 
\textbf{PushBox}\\ 
Behavior Cloning (100)          & 60 & & 4000 & -- & -- \\
HG-Dagger~\cite{kelly2019hg}    & 72 & & 5382 & 39 & 13 \\
IWR~\cite{mandlekar2020human}   & \mathbf{78} & & 4765 & 38 & 5 \\
CEILing (Ours)                  & \mathbf{78} & & 5604 & 41 & \quad\enspace\ 16 \ / \ 1 \\
\midrule 
\textbf{PickupCube}\\ 
Behavior Cloning (100)          & 28 & & 4000 & -- & -- \\
HG-Dagger~\cite{kelly2019hg}    & 90 & & 5054 & 36 & 5 \\
IWR~\cite{mandlekar2020human}   & 86 & & 5198 & 39 & 7 \\
CEILing (Ours)                  & \mathbf{98} & & 4366 & 36 & \quad\enspace\ 3  \ / \ 1 \\
\midrule 
\textbf{PullPlug}\\ 
Behavior Cloning (100)          & 58 & & 4000 & -- & -- \\
HG-Dagger~\cite{kelly2019hg}    & 74 & & 5405 & 39 & 14 \\
IWR~\cite{mandlekar2020human}   & 78 & & 4714 & 33 & 5 \\
CEILing (Ours)                  & \mathbf{80} & & 5799 & 41 & \quad\enspace\ 7  \ / \ 2 \\
\bottomrule
\end{tabularx}

\caption{Results from the real robot experiments. Each policy was trained for $100$ episodes. The feedback rate represents the percentage of steps the human teacher had to intervene with corrective feedback. For CEILing, we also show the negative feedback rate, i.e. the percentage of steps the human teacher provided a negative evaluative feedback. The duration is the amount of real time the policy were trained for and update steps are the number of gradient descent updates of the network. The success rate is the percentage of successful episodes during evaluation.}
\label{tab:table_kuka}
% \vspace{-0.4cm}
\end{table}

In this section, we demonstrate that CEILing is directly applicable to real robot settings. Similar to our simulation experiments, we evaluate on a robot manipulator with 7 degrees of freedom and a wrist mounted camera. We train and evaluate on four different tasks depicted in \figref{fig:kuka_tasks}: Reach, PushBox, PickupCube, PullPlug. We compare the four main baselines: Behavior Cloning (100), HG-Dagger, IWR and CEILing. We did not include the Behavior Cloning (10) and the evaluative feedback baselines because they performed too poorly and were unsafe to evaluate on the real robot. We train each policy for 100 episodes and evaluate it for 50 episodes. The results from these experiments are presented in \tabref{tab:table_kuka}. We observe that CEILing achieves high success rates surpassing the \( 75\% \) threshold across all tasks in the real robot setting as well. The highest correction rate required across tasks was 16\%. This highlights that providing feedback in an interactive learning setting requires much less effort from the human teacher compared to collecting expert demonstrations. It is interesting to note that all the training runs lasted not more than 41 minutes, which also includes the environment resetting. This demonstrates that with CEILing, it is feasible to train manipulation policies within an hour of real-world training, in contrast to multiple hours or days needed by standard reinforcement learning algorithms. Finally, we also evaluate the generalization capabilities of CEILing on the Reach and PushBox tasks. With these experiments, best appreciated in the accompanying video, we show that the learned policies are robust against external disturbances, which were not present during training.

\section{Conclusion}
\label{sec:conclusion}

In this work, we proposed CEILing, an interactive learning framework to address complex robotic manipulation tasks. Compared to standard imitation learning, an interactive learning approach requires much less effort from the human teacher while simultaneously mitigating compounding errors and data mismatch issues. We compared our approach with behavior cloning and different interactive learning strategies, showing competitive performance consistently. In fact, we demonstrated that the best performance can be achieved by trading off a combination of both corrective and evaluative feedback from a human teacher. We presented experimental results on both simulated as well as real-world environments and demonstrated that CEILing can learn complex manipulation policies from visual observations in less than one hour of real-world training. In light of these results, we believe that CEILing and interactive learning in general represent a promising approach to tackle real-world robot learning problems.

%%%%%%%%%%%%%%%%%%%%%%%%%%%%%%%%%%%%%%%%%%%%%%%%%%%%%%%%%%%%%%%%%%%%%%%%%%%%%%%%

\section*{Acknowledgement}

This work was funded by the BrainLinks-BrainTools center of the University of Freiburg.

%%%%%%%%%%%%%%%%%%%%%%%%%%%%%%%%%%%%%%%%%%%%%%%%%%%%%%%%%%%%%%%%%%%%%%%%%%%%%%%%

\footnotesize
\bibliographystyle{IEEEtran}
\bibliography{chisari21ral.bib}

\end{document}